# Building Better Nurse Scheduling Algorithms

Dr Uwe Aickelin

School of Computer Science

University of Nottingham   NG8 1BB   UK

Email: uxa@cs.nott.ac.uk

Tel: +44 (0)11595 14215

Dr Paul White

University of the West of England

Bristol, BS16 1QY, UK

Email: paul.white@uwe.ac.uk

Tel: +44 (0)117 344 3145

Authors are in alphabetical order

**Abstract**

The aim of this research is twofold: Firstly, to model and solve a complex nurse scheduling problem with an integer programming formulation and evolutionary algorithms. Secondly, to detail a novel statistical method of comparing and hence building better scheduling algorithms by identifying successful algorithm modifications. The comparison method captures the results of algorithms in a single figure that can then be compared using traditional statistical techniques. Thus, the proposed method of comparing algorithms is an objective procedure designed to assist in the process of improving an algorithm. This is achieved even when some results are non-numeric or missing due to infeasibility. The final algorithm outperforms all previous evolutionary algorithms, which relied on human expertise for modification.

**Keywords:** Nurse scheduling, evolutionary algorithms, integer programming, statistical comparison method.

In previous work (Aickelin and Dowsland (2000)), satisfactory solutions to a nurse scheduling problem using various evolutionary algorithm approaches had been found. In common with other researchers in this area, those algorithms were built component by component using mainly experience and intuition to identify successful and unsuccessful parts and parameters. Sometimes, it was possible to identify clearly better implementations, e.g. a radically different type of crossover, by comparing the mean performance of the algorithm and the quality of the solutions it produced before and after the changes had been made. However, often differences were subtle with improvements on some data instances and worse results on others. This has been particularly true for finding good parameters such as the mutation rate where extensive domain and problem knowledge was necessary to identify better values.

Additionally there was the problem of some algorithms failing to find feasible solutions for some data instances. Although penalty function approaches (Michalewicz (1995)) are suitable to assign fitness function values *during* the run of the algorithm, their use to compare the *final* best solution of one algorithm to that of another is problematic. Setting good penalty weights is a non-trivial task and often no good single value exists to cover all eventualities. For instance, depending on the weight setting an algorithm that produces infeasible but otherwise low cost solutions might then be preferred over an algorithm that finds feasible solutions. What about comparing two algorithms over a large number of data instances where one found a high number of low quality feasible solutions and the other fewer feasible, but higher quality solutions? In the past (Aickelin and Dowsland (2000) and Aickelin (2002)) a fixed cost (censored cost) had been assigned to any infeasible solution for comparison's sake. However, this raised the question of what level this should be at and whether one infeasible solution might be better than another.

Here we will present a method of statistically comparing algorithms. The implementation is such that any two versions of our algorithms can be compared and better components or parameter values are identified easily. We will use our nurse scheduling problem as a test bed to explain the workings and to show the effectiveness. Without any experience or relying on intuition, we are able to exceed all previous results by building the best heuristic algorithm to solve our problem to date. The proposed method of comparing algorithms is an objective procedure designed to assist in the process of improving an algorithm. This method has been used to build an improved nurse scheduling algorithm. Other innovations could be employed in an attempt to improve the current algorithm e.g. the column generation approach, and the effect of any additional innovation could be assessed using the method outlined. Thus, the proposed method of analysis is not a methodological search of all possible algorithms or improvements, but can assist in the intelligent improvement of parameter estimates and assess the effect of innovative alterations to an algorithm.

Moreover, one should note that the comparison methodology used is not restricted to the nurse scheduling problem discussed here or even to staff scheduling problems in general and can be used to compare any two algorithms solving the same problem. In particular, it was important to us to construct a comparison method that can cope with missing results due to the failure to reach a feasible solution.

The remainder of this paper is organised as follows. The next section introduces the nurse scheduling problem and states the equivalent integer programming formulation. Section 2 briefly explains Genetic Algorithms and show how they can be used to solve the nurse scheduling problem. The subsequent section then justifies the need and defines the statistical comparison algorithm. In Section 4, we use this comparison method to identify successful modifications to the original implementations to build the final algorithm. The paper concludes with a discussion and conclusions of our findings.

## 1 THE NURSE SCHEDULING PROBLEM

Creating good schedules for nurses in today's heavily constrained hospitals is no trivial task. A general overview of nurse scheduling algorithms can be found in Hung (1995), Sitompul and Randhawa (1990) and Bradley and Martin (1990). The rostering problem tackled in this paper can be described as follows. The task is to create weekly schedules for wards of up to 30 nurses by assigning one of a number of possible shift patterns to each nurse. These schedules have to satisfy working contracts and meet the demand for a given number of nurses of different grades on each shift, while being seen to be fair by the staff concerned. The latter objective is achieved by meeting as many of the nurses' requests as possible and considering historical information to ensure that unsatisfied requests and unpopular shifts are evenly distributed.

The problem is complicated by the fact that higher qualified nurses can substitute less qualified nurses but not vice versa. Thus scheduling the different grades independently is not possible. Furthermore, the problem has a special day-night structure as most of the nurses are contracted to work either days or nights in one week but not both. However due to working contracts, the number of days worked is not usually the same as the number of nights. Therefore, it becomes important to schedule the 'right' nurses onto days and nights respectively. These characteristics make this problem challenging for any local search algorithm as finding and maintaining feasible solutions is extremely difficult. Furthermore, due to this special structure previous nurse scheduling algorithms suggested in the literature cannot be used.

As described in detail in Dowsland and Thompson (2000) the problem can be decomposed into three independent stages. The first stage ensures that there are enough nurses to provide adequate cover. The second stage assigns the nurses to the correct number of day or night shifts. A final phase allocates those working on a particular day to the early or late shift on that day. Phases 1 and 3 can be solved using classical optimisation models. Thus, this paper deals with the highly constrained second step.

The numbers of days or nights to be worked by each nurse defines the set of feasible weekly work patterns for that nurse. These will be referred to as shift patterns or shift pattern vectors in the following. For example (1111100 0000000) would be a pattern where the nurse works the first 5 days and no nights. Then in phase 3, these days would be split into early and late shifts. Depending on their contacts, some nurses can work either days or nights or combinations of both. For each nurse, $i$, and each shift pattern, $j$, all the information concerning the desirability of that pattern for that nurse is captured in a single numeric preference cost $p_{ij}$. This was done in close consultation with the hospital and is a weighted sum of the following factors: Basic shift-pattern cost, general day / night preferences, specific requests, continuity problems, rotating nights / weekends and other working history information. Further details can be found in Dowsland (1998).

The problem can be formulated as an integer linear program as follows.

Indices:

$i = 1, \ldots n$ nurse index.

$j = 1, \ldots m$ shift pattern index.

$k = 1, \ldots 14$ day and night index (1, ... 7 are days and 8, ... 14 are nights).

$s = 1, \ldots p$ grade index.

Decision variables:

$$x_{ij} = \begin{cases} 1 & \text{nurse } i \text{ works shift pattern } j \\ 0 & \text{else} \end{cases}$$

Parameters:

$n$ = Number of nurses.

$m$ = Number of shift patterns.

$p$ = Number of grades.

$$a_{jk} = \begin{cases} 1 & \text{shift pattern } j \text{ covers day / night } k \\ 0 & \text{else} \end{cases}$$

$$q_{is} = \begin{cases} 1 & \text{nurse } i \text{ is of grade } s \text{ or higher} \\ 0 & \text{else} \end{cases}$$

$p_{ij}$ = Preference cost of nurse $i$ working shift pattern $j$.

$N_i$ = The number of working shifts per week of nurse $i$ if night shifts are worked.

$D_i$ = The number of working shifts per week of nurse $i$ if day shifts are worked.

$B_i$ = The number of working shifts per week of nurse $i$ if both day and night shifts are worked.

$R_{ks}$ = Demand of nurses with grade $s$ on day respectively night $k$.

$F(i)$ = Set of feasible shift patterns for nurse $i$, where $F(i)$ is defined as

$$F(i) = \begin{cases} \sum_{k=1}^{7} a_{jk} = D_i & \forall j \in \text{day shifts} \\ \text{or} \\ \sum_{k=8}^{14} a_{jk} = N_i & \forall j \in \text{night shifts} \\ \text{or} \\ \sum_{k=1}^{14} a_{jk} = B_i & \forall j \in \text{combined shifts} \end{cases} \quad \forall i$$

Objective function:

$$\sum_{i=1}^{n} \sum_{j \in F(i)}^{m} p_{ij} x_{ij} \rightarrow \text{min!}$$

Subject to:

1. Every nurse works exactly one feasible shift pattern:

$$\sum_{j \in F(i)} x_{ij} = 1 \qquad \forall i \qquad (I)$$

2. The demand for nurses is fulfilled for every grade on every day and night:

$$\sum_{j \in F(i)} \sum_{i=1}^{n} q_{is} a_{jk} x_{ij} \geq R_{ks} \quad \forall k, s \qquad (II)$$

Constraint set (I) ensures that every nurse works exactly one shift pattern from his/her feasible set, and constraint set (II) ensures that the demand for nurses is covered for every grade on every day and night. Note that the definition of $q_{is}$ is such that higher graded nurses can substitute those at lower grades if necessary. Typical problem dimensions are 30 nurses of three grades and 411 shift patterns. Thus, the IP formulation has about 12000 binary variables and 100 constraints. Although this is only a moderately sized problem, Fuller (1998) shows that some problem instances remain unsolved after more than 12 hours of computation time on a Pentium II PC using professional software.

## 2   GENETIC ALGORITHM IMPLEMENTATION

In recent years, Genetic Algorithms (GAs) have become increasingly popular for solving complex optimisation problems such as those found in the areas of scheduling or timetabling. GAs are generally attributed to Holland (1976) and his students in the 1970s, although evolutionary computation dates back further (refer to Fogel (1998) for an extensive review of early approaches). GAs are stochastic meta-heuristics that model some features of natural evolution Canonical GAs were not intended for function optimisation, as discussed by De Jong (1993). However, slightly modified versions proved very successful. For an introduction to GAs for function optimisation, see Deb (1996). Many examples of successful implementations can be found in Bäck (1993), Chaiyaratana and Zalzala (1997) and others.

In a nutshell, GAs mimic the evolutionary process and use the idea of the survival of the fittest. Starting with a population of randomly created solutions, better ones are more likely to be chosen for recombination into new solutions, i.e. the fitter a solution, the more likely it is to pass on its information to future generations of solutions. In addition to recombining solutions, new solutions may be formed through mutating or randomly changing old solutions. Some of the best solutions of each generation are kept whilst the others are replaced by the newly formed solutions. The process is repeated until stopping criteria are met.

However, constrained optimisation with GAs remains difficult. The root of the problem is that simply following the building block hypothesis, i.e. combining good building blocks or partial solutions to form good full solutions, is no longer enough, as this does not check for constraint consistency. To solve this dilemma, many ideas have been proposed of which the major ones are penalty functions and repair. A good overview of these and most other techniques can be found in Michalewicz (1995).

Here we use two approaches: First, an encoding that follows directly from the Integer Programming formulation. Each individual represents a full one-week schedule, i.e. it is a string of $n$ elements with $n$ being the number of nurses. The *ith* element of the string is the index of the shift pattern worked by nurse $i$. For example, if we have 5 nurses, the string (1, 17, 56, 67, 3) represents the schedule in which nurse 1 works pattern 1, nurse 2 pattern 17, and so on. Full details of this algorithm can be found in Aickelin and Dowsland (2000).

The second approach presented here is the combination of an indirect GA with a separate heuristic decoder function. Here, the GA tries to find the best possible ordering of the nurses, which is then fed into the greedy decoder that assembles the actual solution. One way of looking at this decoder is as an extended fitness function calculation, i.e. the decoder determines the fitness of a solution after it has built a schedule from the permutation of nurses. For instance, a very simple (and perhaps not very good) decoder could take the nurses in the order decided by the GA and then assign each one her lowest cost shift pattern. Another simple decoder might take the nurses in order and assign each one to the currently most uncovered shift, taking into account nurses already scheduled. The actual decoder used considers a number of objectives, such as cost, nurses' preferences and cover, simultaneously and tries to find the best shift pattern for each nurse based on all objectives. The GA and the decoders used are fully described to solve a related multiple-choice problem in Aickelin and Dowsland (2002). The advantages of this approach are that all problem specific information is contained within the decoder, whilst the GA can be kept canonical. The only difference from a standard GA is the need for permutation-based crossover and mutation operators as explained for instance in Goldberg (1989).

For both algorithms, the fitness of completed solutions has to be calculated. Unfortunately, feasibility cannot be guaranteed for either method, as otherwise an unlimited supply of nurses would be necessary. Therefore, we need a penalty function approach. Since the chosen encoding automatically satisfies constraint set (I) of the integer programming formulation, we can use the following formula, where $w_{demand}$ is the penalty weight, to calculate the fitness of solutions. Hence the penalty is proportional to the number of uncovered shifts and the fitness of a solution is calculated as follows.

$$\sum_{i=1}^{n}\sum_{j=1}^{m}p_{ij}x_{ij} + w_{demand}\sum_{k=1}^{14}\sum_{s=1}^{p}\max\left[R_{ks} - \sum_{i=1}^{n}\sum_{j=1}^{m}q_{is}a_{jk}x_{ij}; 0\right] \rightarrow \min!$$

## 3  THE NEED FOR A COMPARISON ALGORITHM

As outlined in the previous section, the GA might build infeasible solutions trying to solve the problem and no polynomial time repair algorithm for this class of problem is known. Even worse, because of this the only way to compare feasible and infeasible solutions within a GA is a penalty based method and hence some infeasible solution will be 'fitter' than feasible solutions and might dominate. Although some of this might be countered with high penalty weights, past work (Aickelin and Dowsland (2000)) has shown that if the weights are set too high often no good solutions are found at all. Furthermore, for this particular nurse scheduling problem it is known that for some data instances only very few feasible solutions exist in a large solution space. Thus, the GA will have to find the needle in the haystack. Consequently, there will be algorithm runs that terminate without ever encountering a feasible solution and hence their best solution will have a value of 'infeasible'.

Clearly, this is undesirable and better algorithms should be less and less likely to end up with an infeasible solution. However, the question remains how to build such well performing algorithms. The standard way of achieving this is to build a simple algorithm, assess how it performs, then add a new feature and compare the results. If the results are improved, the new feature is kept otherwise, it is discarded. This is then repeated for a number of features until the algorithm shows the desired performance. In a similar fashion, parameters can be tuned for better results.

Due to the reliability of heuristics on random numbers fair comparisons are only possible if the same problem instance is solved multiple times to avoid bias. We use 20 runs per problem instance. Thus, when deciding if an improvement is significant, it is traditionally the mean results before and after introducing the new feature that are compared. Alternatively, the best or worst cases or quartiles can be used as a basis for comparison. However, the question remains how to deal with algorithms that terminate with an infeasible solution as their best result. In these situations, the mean of the results can no longer be calculated, but one might argue that this could be substituted by the median. However, as the following examples show, this might not prove to be sufficient.

For instance, suppose we have two algorithms, say ALG1 and ALG2. Further suppose 10 trials of each algorithm were tried on the same problem instance. Let the values of the objective functions obtained for ALG1 be "infeasible" for the first six trials and be costs of 1, 2, 3 and 4 for the other four trials. Likewise, let the solutions obtained for ALG2 for be "infeasible" for the first 6 trials and let the cost for the next four trials be 1, 6, 7 and 8. Let us assume that as in the nurse scheduling problem lower costs are preferred. In this case, both algorithms have the same median solution ("infeasible") and both algorithms have the same cost for their best solution (cost of 1). What would be required is a comparison procedure that would globally rank ALG1 to be better than ALG2.

Although looking at quartiles might have worked for the simple example above, for a more realistic situation comparisons are even more difficult. Figure 1 plots the best solution found for three variants of our algorithms V2, V4 and V6 for the 52 weekly scheduling problems. Inspection of these best costs would tend to suggest that algorithm V6 is inferior to algorithms V2 and V4, and the differences between V4 and V6 are so small that the two algorithms may be viewed as being equally effective. To a certain extent the same conclusions would be drawn if attention was focussed on the median solution obtained (see Figure 2). However, if attention is restricted to the upper quartile or to the worst solution obtained then it is noticeable that V4 tends to be superior to V2 (see Figure 3 and Figure 4). So which one is best?

What is needed is a comparison procedure that is sensitive to changes in the location of a distribution but which also depends upon other distribution properties. Let us now construct such a method. Consider the general case where an algorithm $i$ is to be compared with an algorithm $j$ on a variant $p$ of a problem. Suppose that $K$ runs for algorithm $i$ are to be performed with each run differing in initial starting conditions. Likewise, for the same variant of the problem suppose that $L$ runs of algorithm $j$ are to be performed. Let $c_{i,k,p}$ denote the cost obtained for run $k$ of algorithm $i$ on problem $p$. Likewise let $c_{j,l,p}$ denote the cost obtained for run $l$ of algorithm $j$ on problem $p$. If $c_{i,k,p} < c_{j,l,p}$ then the solution for algorithm $i$ on run $k$ may be viewed as being better than the solution obtained by algorithm $j$ on trial $l$.

Note that the "cost" associated with the solution need not be stated explicitly. For instance, suppose that a feasible solution on trial $l$ for algorithm $j$ was not found ($c_{j,l,p}$ = "infeasible") and for the same problem suppose a feasible solution on trial $k$ for algorithm $i$ was found. In many cases of this nature, we would want to interpret this example situation as $c_{i,k,p} < c_{j,l,p}$.

Suppose $K$ trials of algorithm $i$ and $L$ trials of algorithm $j$ on variant $p$ of a problem are obtained with associated costs $c_{i,k,p}$ and $c_{j,l,p}$ ($k = 1, \ldots K; l = 1, \ldots L$). Consider

$$D_{i,j,k,l,p} = \begin{cases} +1 & \text{if } c_{i,k,p} < c_{j,l,p} \\ 0 & else \\ -1 & \text{if } c_{i,k,p} > c_{j,l,p} \end{cases}$$

Thus, $D_{i,j,k,l,p}$ indicates whether a particular instance of algorithm $i$ is "better", "the same" or "worse" than algorithm $j$. An aid to assessing whether algorithm $i$ is typically better than algorithm $j$ on problem $p$ is to consider the average value of $D_{i,j,k,l,p}$ over all possible pairwise comparisons i.e.,

$$E_{i,j,p} = \sum_{k=1}^{K} \sum_{l=1}^{L} D_{i,j,k,l,p} \bigg/ LK \qquad (1)$$

Clearly, $E_{i,j,p} = -E_{j,i,p}$ and $-1 \leq E_{i,j,p} \leq +1$ with the extreme values ($\pm 1$) being obtained if one algorithm is "better" than the other in every observed instance. Values of $E_{i,j,p}$ equal to zero would occur when there is no difference between the algorithms or when the pairwise comparison frequency for "better" is equal to the pairwise comparison frequency for "worse".

In addition, if $0 < E_{h,i,p}$ and $0 < E_{i,j,p}$ then $0 < E_{h,j,p}$ (i.e., if on problem $p$ algorithm $h$ is typically better than algorithm $i$ and algorithm $i$ is typically better than algorithm $j$ then algorithm $h$ is also "better than" to $j$). Hence, $0 < E_{i,j,p}$ permits a ranking for the algorithms $h$, $i$ and $j$ for problem $p$.

To illustrate suppose that a particular algorithm (ALG3) is used ten times to solve a particular problem. Let the costs for the ten solutions be 1, 1, 1, 2, 3, 3, 3, 4, 5, and 5. Let the ten costs from ten trials from a second algorithm (ALG4) be 2, 4, 5, 5, 6, 7, 8, 8, "infeasible" and "infeasible". Likewise let the ten costs for ten trials using a third algorithm (ALG5) be 3, 4, 5, 8, 9, 10, 10, "infeasible", "infeasible" and "infeasible". Inspection of these sample data would strongly suggest that ALG3 is "better" than ALG4, and that ALG4 is "better" than ALG5. Indeed $0 < E_{ALG3,ALG4,p} = +0.78$ (i.e., 86 comparisons of ALG3 with ALG4 yield "+1", 6 comparisons yield "0" and 8 comparisons yield "–1"). Further, $E_{ALG3,ALG5,p} = +0.84$ and $E_{ALG4,ALG5,p} = +0.33$, and hence under this metric we may establish the simple rank ordering of ALG3 < ALG4 < ALG5 where "<" denoting "less than", may in this context be viewed as "better than".

If the hypothetical algorithms ALG3, ALG4 and ALG5 were applied to a variation of the problem then it is possible that a different rank ordering would be established. That is to say, the rank ordering could vary from one variation of a problem to another. In these instances, interest may focus on whether the rank orderings from problem to problem may be viewed as being "random" (meaning each permutation of rank orderings is equiprobable) or whether there is statistical evidence of some structure in the rank orderings over and above that anticipated by chance. In these cases, Friedman's test may be applied with the algorithms being viewed as levels of a single factor and the blocks being the variants of the problem under consideration. For an explanation of Friedman's test, see Lehmann (1975) or Conover (1980). If there is statistical evidence of some systematic arrangement of the relative rank ordering of the algorithms then scope exists to determine which algorithms differ. This can be achieved either by pairwise application of the Sign test based on the rank positions (i.e., the two-sample equivalent of Friedman's test) or by pairwise application of Wilcoxon's test using $E_{i,j,p}$ $(p = 1, ... P)$ as the difference measure between algorithm $i$ and $j$ on problem $p$. In applying Wilcoxon's test the statistical null hypothesis being tested would be for $E_{i,j,p}$ $(p = 1, ... P)$ to be independent realisations from a symmetric distribution centred on zero. For an explanation of Wilcoxon's Signed Rank Sum Test see Lehmann (1975) or Conover (1980).

## 4   ANALYSIS AND DISCUSSION

For the problem under investigation eight algorithms were initially compared (say *V1, V2, V3, V4, V5, V6, V7, V8*). Brief descriptions of these eight algorithms are given in Table 1. These particular eight algorithms were chosen as they represented milestones in our original research (Aickelin and Dowsland (2000) and (2002)). Whilst building those algorithms we constantly had to decide which features to keep or what parameter values to set. Typical questions were "Does adding a hillclimber improve results?", "Which algorithm is better, the direct or the indirect one?" or "What crossover should we use?". To answer these questions, we looked at the results obtained, but were faced with the problems outlined in the beginning of the previous section. Ultimately, we had to rely on our expertise to build the 'perfect' algorithm. After studying the problem for 3 years, the best we could build was V4.

In this section, we will show that with the new comparison measure we can improve upon our results because we no longer have to rely on human expertise to decide if a component or parameter setting is successful. First, we confirm that the comparison method works by ranking the milestone algorithms, for which we 'know' the rank positions. Then we will revisit some algorithm components and parameter settings where we had problems deciding the merit of one over another. During these tests we will show that the comparison measure is both sensitive enough to pick up improvements that we had missed previously, whilst being robust enough not to show a preference when no significant improvement had been made.

In order to have representative data, the weekly rostering requirements was obtained for a full calendar year, i.e. for 52 weeks. During all tests all 52 variants of the scheduling problem were considered (i.e., $p = 1, ... 52$) with each variant being the rostering of nurses for each of the 52 weeks. Twenty trials for each algorithm on each variant $p$ were performed. Application of equation 1 to the costs obtained from the trial solutions for each week permitted a rank ordering of the algorithms for each week, i.e., 52 rank orderings of the algorithms. An extract of the data for the first three weeks, the associated values for $E_{Vi,Vj,1}$ and the associated rank orderings are given in Tables 2, 3 and 4 respectively.

Statistically significant differences for the algorithms were found using Friedman's test ($S$ = 290.54, *df* = 7, *p-value* < 0.001), where $S$ is the value of Friedman's statistic on 7 degrees of freedom (*df*) and *p-value* denotes the p-value used for the test. The average rank position of each algorithm provides prima facie evidence of an overall ranking of the algorithms with V3 < V6 < V8 < V7 < V2 < V1 < V5 < V4 (i.e., algorithm V4 is tentatively regarded as the best, V5 the second best and so on through to V3 and V6 which are tentatively regarded as the worst). For illustration purposes Table 5 gives $E_{V6,V3,1}$ *(p = 1, ..., 52)* for algorithms V6 and V3 and Table 6 gives $E_{V6,V8,1}$ *(p = 1, ..., 52)*.

To investigate whether there is statistical evidence of a difference between algorithm V6 and algorithm V3 one approach would be to use the Wilcoxon Signed Rank test to test whether the sample values $E_{V6,V8,1}$ *(p = 1, ... 52)* can be thought of as random sample from a symmetric population centred on zero. Another approach would be to test whether the signs of the sample values are considered to come from independent Bernoulli distributions with parameter $\pi$ = 0.5 (i.e. a negative value just as likely as a positive value after ignoring any values equal to 0.). Application of the Wilcoxon test to the data in Table 5 confirms that there is no statistical evidence of a difference between algorithm V6 and algorithm V3 ($T_+$ = 760, $T_-$ = 566, Z = 0.909, n = 51, p-value > 0.35, two-sided test). Similarly, note for the sample data in Table 5, there are 30 positive instances of $E_{V6,V3,p}$, 21 negative instances of $E_{V6,V3,p}$ and 1 instance equal to zero. Application of the Sign test to these frequencies fails to confirm a statistically significant difference between V6 and V3 (B = 30, n = 51, p-value > 0.25, two-sided).

Application of the Wilcoxon test to the data in Table 6 shows that algorithm V8 is "better than" V6 ($T_+$ = 1267.5, $T_-$ = 58.5, Z = 5.67, n = 51, p < 0.001, two-sided test). Similarly for the sample data in Table 6, there are 45 instances of $E_{V6,V8,p}$ which are positive, 6 instances of $E_{V6,V8,p}$ are negative, and 1 instance equal to zero. Application of the Sign test to these frequencies confirms a statistically significant difference between V8 and V6 (B = 45, n = 51, p-value < 0.001, two-sided).

Application of Wilcoxon's Rank Sum Test and/or the Sign test in the above manner leads to the conclusion that V4 is significantly better than V5. Repeated applications of these tests lead to the conclusion that V5 is significantly better than V1, V1 is significantly better than V2, V2 is significantly better than V7, V7 is significantly better than V8 and V8 is significantly better than V6. The data do not provide statistical evidence of a difference between V6 and V3.

The above conclusions are based on counting the number of instances one algorithm outperforms another. In the counting process, the magnitude of any difference is not explicitly accounted for. Indeed the magnitude of a difference may be difficult to quantify if a feasible solution is not obtained. For these reasons we may consider $D_{i,j,k,l,p}(\alpha)$ defined by:

$$D_{i,j,k,l,p}(\alpha) = \begin{cases} +1 & \text{if } c_{i,k,p} < c_{j,l,p} = \text{"infeasible"} \\ +\alpha & \text{if } c_{i,k,p} < c_{j,l,p} \\ 0 & \text{else} \\ -\alpha & \text{if } c_{i,k,p} > c_{j,l,p} \\ -1 & \text{if "infeasible"} = c_{i,k,p} > c_{j,l,p} \end{cases}$$ where $\alpha$ is a parameter subject to $0 \leq \alpha \leq 1$.

The earlier conclusions for the ranking of the algorithms V1, V2, V3, V4, V5, V6, V7 and V8 where obtained for the special case of $D_{i,j,k,l,p}(\alpha) = D_{i,j,k,l,p}(1)$. In fact, repeat analyses using α = 0.5, 0.6, 0.7, 0.8 and 0.9 leads to the same broad conclusions as when α = 1.0.

Having shown that the comparison method works, let us now turn our attention to algorithm features that we had previously discarded as we could not perceive an improvement whilst using them. Hence, a second set of algorithms U1, U2, U3, U4, U5, U6, U7 and U8 are compared. A description of these algorithms is given in Table 1. Each of these algorithms was used 20 times on each of the 52 weekly scheduling problems and the above process was applied. In summary the poorest performing algorithms were U4 and U6 (no statistical evidence that they differ), followed by U3 which was significantly better than U4 and U6. Algorithms U2 and U8 were identical and significantly better than U3. There was no statistical evidence of a difference between U1 and U5 although both showed a statistically significant improvement over algorithms U2 and U8. Finally, algorithm U7 showed a statistically significant improvement over U1 and U5 and was considered the best from this set of algorithms.

Thus, we stand corrected from our previous result built upon our expertise. Figure 5 shows summary results for U7 and V4 (best indirect GA and our previous champion) and V7 (the best direct GA). The worst solution for each problem instance from U7 was compared with the worst solution found for each problem instance using U8 (which equals V4). In this comparison U7 obtained the better "worse" solution on 22 instances, U8 obtained the better "worse" solution on 14 instances and the same "worse" solution is obtained in 16 instances. If the median solutions are compared then U8 obtains the better median solution on 23 instances, U7 obtains the better median solution on 5 instances and the same median solution is obtained on 24 instances. If the best solutions are compared then U7 obtains the better solution on 9 instances, U8 obtains the better solution on 4 instances and the same best solution is obtained on 39 instances. It is worth pointing out in one instance a feasible solution was not obtained by U7 on one occasion whereas the best solution found by U8 was always feasible.

To confirm that the comparison method is not too sensitive we decided to use a third set of algorithms (W1, W2, W3, W4, W5, W6, W7 and W8) on the scheduling problem in exactly the same way as before. A description of these algorithms is given in Table 1. The algorithms were chosen such that we were reasonably convinced that none was much better than any of the others. Prima facie evidence might suggest that algorithm W4 is marginally better than the others (with W5, W8, W7 and W6 being identical to each other and marginally better than W1, W2 and W3). However, we would expect the comparison algorithm to be robust enough not to show any statistically significant difference. This was confirmed, as the observed differences by our method do not provide statistical evidence to claim a significant difference amongst the algorithms in this set.

# 5    CONCLUSIONS

The pairwise comparison procedure outlined provides a method of ranking algorithms on a problem instance that is applicable when some solutions are infeasible. In the presence of "infeasible" some sample statistics such as the mean or the standard deviation cannot be legitimately calculated unless it is justifiable to replace "infeasible" with a known finite cost. In summary, the comparison method condenses the results of algorithms to a single figure that can then be compared using traditional statistical techniques. This is achieved even when some results are "missing" due to infeasibility.

Focussing on the best solution alone is wrong, particularly in the algorithm building and refinement stage. Take for example two hypothetical algorithms ALG 6 and ALG7. Suppose ALG6 in 20 trials yields 19 infeasible solutions and one solution with a cost of zero. Further suppose that ALG7 finds 20 feasible solutions with costs greater than zero. Is it better to improve ALG6 or ALG7? If we focus on best outcomes only, we neglect substantial improvements that might have happened for the majority of solutions. On the other hand, if we differentiate the results based on the whole distribution we can more reliably spot new components or parameter settings that improve results. Such a method is provided by our comparison algorithm and it works on a real life problem as the improved results for the nurse scheduling problem show.

| Algorithm | (In)Direct GA | Bound | Crossover | Elitism | Extras | Rank |
|---|---|---|---|---|---|---|
| V1 | Indirect | Look-ahead | 80% uniform | 10% | --- | 3 |
| V2 | Indirect | None | 80% uniform | 10% | --- | 4 |
| V3 | Indirect | Greedy | 80% uniform | 10% | --- | 7.5 |
| **V4** | **Indirect** | **Look-ahead** | **Automatic** | **10%** | **---** | **1** |
| V5 | Indirect | Look-ahead | 1-point | 10% | --- | 2 |
| V6 | Direct | None | 80% uniform | 10% | --- | 7.5 |
| V7 | Direct | None | 80% uniform | 10% | Sub-Populations | 5 |
| V8 | Direct | None | 80% uniform | 10% | Hillclimber | 6 |
| U1 | Indirect | Look-ahead | 50% uniform | 10% | --- | 2.5 |
| U2 | Indirect | Look-ahead | 67% uniform | 10% | --- | 4.5 |
| U3 | Indirect | Look-ahead | 75% uniform | 10% | --- | 6 |
| U4 | Indirect | Look-ahead | 80% uniform | 10% | --- | 7.5 |
| U5 | Indirect | Look-ahead | Automatic | 10% | Auto-weights | 2.5 |
| U6 | Indirect | Look-ahead | Automatic | 10% | --- | 7.5 |
| **U7** | **Indirect** | **Look-ahead** | **80% uniform** | **10%** | **Auto-weights** | **1** |
| U8 = V4 | Indirect | Look-ahead | Automatic | 10% | --- | 4.5 |
| W1 | Indirect | Look-ahead | 80% uniform | 50% | --- | 4.5 |
| W2 | Indirect | Look-ahead | 80% uniform | 40% | --- | 4.5 |
| W3 | Indirect | Look-ahead | 80% uniform | 30% | --- | 4.5 |
| W4 | Indirect | Look-ahead | 80% uniform | 20% | --- | 4.5 |
| W5 | Indirect | Look-ahead | 80% uniform | 5% | --- | 4.5 |
| W6 | Indirect | Look-ahead | Rank-based | 10% | --- | 4.5 |
| W7 | Indirect | Look-ahead | Bound-based | 10% | --- | 4.5 |
| W8 = V4 | Indirect | Look-ahead | Automatic | 10% | --- | 4.5 |

Table 1: Details of the GAs compared in the paper. The bold entries represent the best one of each set, note that for set W the differences were not statistically significant. The rank column gives the rank position of an algorithm relative to others in its set as determined by our comparison method. Full descriptions of these algorithms can be found in Aickelin and Dowsland (2000) and (2002). Briefly, (In)direct refers to the type of algorithm used as described in section 2 of the paper; Bound refers to how intelligently the solutions are built (i.e. not applicable to the direct version); Crossover gives the type of crossover used (automatic meaning the algorithm tries to decide itself), Elitism gives the percentage of the best solution carried over from one generation to the next; Auto-weights indicates that the algorithm tries to optimise some further parameters itself.

| Week | V1 | V2 | V3 | V4 | V5 | V6 | V7 | V8 |
|---|---|---|---|---|---|---|---|---|
| 1 | 00, 00, | 00, 00, | 123, 127, | 00, 00, | 00, 00, | 01, 03, | 00, 00, | 00, 00, |
|   | 00, 00, | 00, 00, | 129, NS, | 00, 00, | 00, 00, | 05, 05, | 00, 00, | 00, 00, |
|   | 00, 00, | 00, 00, | NS, NS, | 00, 00, | 00, 00, | 06, 07, | 00, 00, | 00, 02, |
|   | 00, 00, | 01, 01, | NS, NS, | 00, 00, | 00, 00, | 08, 14, | 00, 01, | 02, 03, |
|   | 00, 00, | 01, 01, | NS, NS, | 00, 00, | 00, 00, | 18, 20, | 01, 01, | 03, 03, |
|   | 00, 00, | 01, 01, | NS, NS, | 00, 00, | 00, 00, | 25, 29, | 02, 02, | 03, 03, |
|   | 00, 00, | 01, 01, | NS, NS, | 00, 00, | 00, 00, | 43, NS, | 02, 03, | 10, 19, |
|   | 00, 00, | 02, 02, | NS, NS, | 00, 00, | 00, 00, | NS, NS, | 03, 03, | 21, 27, |
|   | 00, 00, | 02, 02, | NS, NS, | 00, 00, | 00, 02, | NS, NS, | 03, 03, | 64, 97, |
|   | 00, 00 | 02, 04 | NS, NS | 00, 00 | 02, 02 | NS, NS | 03, 05 | NS, NS |
| 2 | 13, 14, | 19, 19, | 106, 109, | 12, 12, | 12, 13, | 19, 21, | 12, 12, | 12, 12, |
|   | 18, 19, | 20, 20, | 109, 112, | 12, 12, | 20, 20, | 21, 23, | 12, 12, | 12, 12, |
|   | 20, 21, | 20, 21, | 112, 119, | 16, 19, | 20, 21, | 23, 24, | 13, 13, | 12, 13, |
|   | 21, 21, | 21, 21, | 126, 131, | 19, 19, | 21, 21, | 24, 24, | 13, 13, | 13, 13, |
|   | 21, 22, | 21, 22, | 132, 132, | 19, 20, | 21, 21, | 25, 26, | 13, 14, | 14, 14, |
|   | 22, 22, | 22, 22, | 134, 142, | 20, 20, | 21, 22, | 28, 29, | 14, 14, | 15, 16, |
|   | 22, 22, | 22, 22, | 148, 150, | 20, 20, | 22, 22, | 30, 30, | 15, 15, | 19, 19, |
|   | 22, 22, | 23, 23, | 151, 160, | 21, 21, | 22, 22, | 31, 31, | 15, 16, | 20, 20, |
|   | 22, 22, | 23, 25, | 163, 169, | 21, 22, | 22, 23, | 32, 32, | 16, 16, | 21, 21, |
|   | 28, 29 | 28, 31 | 170, 174 | 22, 22 | 23, 24 | 37, 49 | 18, 25 | 22, 31 |
| 3 | 18, 19, | 20, 20, | 119, 141, | 18, 18, | 18, 18, | 03, 19, | 18, 18, | 18, 18, |
|   | 19, 19, | 20, 20, | 147, 157, | 19, 19, | 19, 19, | 20, 21, | 18, 18, | 18, 18, |
|   | 20, 20, | 20, 20, | 158, 159, | 19, 19, | 19, 19, | 24, 25, | 18, 18, | 19, 20, |
|   | 20, 26, | 20, 20, | 159, 159, | 19, 19, | 19, 20, | 28, 28, | 18, 19, | 20, 20, |
|   | 26, 26, | 20, 23, | 160, 161, | 20, 20, | 20, 20, | 28, 29, | 19, 19, | 21, 21, |
|   | 28, 28, | 27, 27, | 164, 167, | 20, 20, | 20, 20, | 30, 32, | 19, 20, | 21, 30, |
|   | 28, 28, | 28, 28, | 168, 174, | 20, 20, | 20, 22, | 32, 34, | 20, 21, | 33, 48, |
|   | 28, 35, | 28, 28, | 178, 194, | 20, 22, | 26, 26, | 36, 36, | 21, 21, | 50, 52, |
|   | 36, 38, | 29, 36, | 205, 208, | 26, 27, | 28, 28, | 40, 42, | 26, 27, | 55, 169, |
|   | 40, 40 | 38, 49 | 211, NS | 28, 28 | 28, 28 | 48, NS | 27, 30 | NS, NS |

Table 2: Costs for 20 runs of algorithms V1, V2, V3, V4, V5, V6, V7 and V8 on weeks 1, 2 and 3. NS stands for no feasible solution obtained.

| Week 1 | $E_{1,2,1} = +0.70$ | $E_{1,3,1} = +1.00$ | $E_{1,4,1} = 0.000$ | $E_{1,5,1} = +0.15$ | $E_{1,6,1} = +1.00$ |
|---|---|---|---|---|---|
|  | $E_{1,7,1} = +0.65$ | $E_{1,8,1} = +0.75$ | $E_{2,3,1} = +1.00$ | $E_{2,4,1} = -0.70$ | $E_{2,5,1} = -0.50$ |
|  | $E_{2,6,1} = +0.95$ | $E_{2,7,1} = +0.17$ | $E_{2,8,1} = +0.52$ | $E_{3,4,1} = -1.00$ | $E_{3,5,1} = -1.00$ |
|  | $E_{3,6,1} = -0.60$ | $E_{3,7,1} = -1.00$ | $E_{3,8,1} = -0.89$ | $E_{4,5,1} = +0.15$ | $E_{4,6,1} = +1.00$ |
|  | $E_{4,7,1} = +0.65$ | $E_{4,8,1} = +0.75$ | $E_{5,6,1} = +0.99$ | $E_{5,7,1} = +0.53$ | $E_{5,8,1} = +0.70$ |
|  | $E_{6,7,1} = -0.92$ | $E_{6,8,1} = -0.47$ | $E_{7,8,1} = +0.40$ |  |  |
| Week 2 | $E_{1,2,2} = +0.13$ | $E_{1,3,2} = +1.00$ | $E_{1,4,2} = -0.50$ | $E_{1,5,2} = -0.05$ | $E_{1,6,2} = +0.71$ |
|  | $E_{1,7,2} = -0.80$ | $E_{1,8,2} = -0.64$ | $E_{2,3,2} = +1.00$ | $E_{2,4,2} = -0.62$ | $E_{2,5,2} = -0.15$ |
|  | $E_{2,6,2} = +0.64$ | $E_{2,7,2} = -0.91$ | $E_{2,8,2} = -0.73$ | $E_{3,4,2} = -1.00$ | $E_{3,5,2} = -1.00$ |
|  | $E_{3,6,2} = -1.00$ | $E_{3,7,2} = -0.89$ | $E_{3,8,2} = -1.00$ | $E_{4,5,2} = +0.52$ | $E_{4,6,2} = +0.89$ |
|  | $E_{4,7,2} = -0.55$ | $E_{4,8,2} = -0.28$ | $E_{5,6,2} = +0.75$ | $E_{5,7,2} = -0.75$ | $E_{5,8,2} = -0.62$ |
|  | $E_{6,7,2} = -0.96$ | $E_{6,8,2} = -0.88$ | $E_{7,8,2} = +0.17$ |  |  |
| Week 3 | $E_{1,2,3} = -0.02$ | $E_{1,3,3} = +1.00$ | $E_{1,4,3} = -0.48$ | $E_{1,5,3} = -0.40$ | $E_{1,6,3} = +0.27$ |
|  | $E_{1,7,3} = -0.56$ | $E_{1,8,3} = +0.06$ | $E_{2,3,3} = +1.00$ | $E_{2,4,3} = -0.55$ | $E_{2,5,3} = -0.44$ |
|  | $E_{2,6,3} = +0.36$ | $E_{2,7,3} = -0.58$ | $E_{2,8,3} = +0.07$ | $E_{3,4,3} = -1.00$ | $E_{3,5,3} = -1.00$ |
|  | $E_{3,6,3} = -0.90$ | $E_{3,7,3} = -1.00$ | $E_{3,8,3} = -0.74$ | $E_{4,5,3} = +0.10$ | $E_{4,6,3} = +0.70$ |
|  | $E_{4,7,3} = -0.17$ | $E_{4,8,3} = +0.34$ | $E_{5,6,3} = +0.65$ | $E_{5,7,3} = -0.25$ | $E_{5,8,3} = +0.29$ |
|  | $E_{6,7,3} = -0.72$ | $E_{6,8,3} = -0.10$ | $E_{7,8,3} = +0.41$ |  |  |

Table 3: $E_{i,j,p}$ (i = 1,…7; j = i,…,8; p = 1,2,3).

|        | V1  | V2 | V3 | V4  | V5 | V6 | V7 | V8 |
|--------|-----|----|----|-----|----|----|----|----|
| Week 1 | 7.5 | 5  | 1  | 7.5 | 6  | 2  | 4  | 3  |
| Week 2 | 4   | 3  | 1  | 6   | 5  | 2  | 8  | 7  |
| Week 3 | 4   | 5  | 1  | 7   | 6  | 2  | 8  | 3  |

Table 4: Rank positions for Weeks 1, 2 and 3. Without any loss of generality, the ranking within each week has been from 1 to 8 with 1 denoting worst and 8 denoting best. Method of mid-ranks used for tied positions.

| (1) +0.5975 | (2) +1.000 | (3) +0.9025 | (4) +1.000 | (5) +0.7500 | (6) -0.0600 | (7) +0.0500 | (8) +0.3375 |
|---|---|---|---|---|---|---|---|
| (9) -0.1875 | (10) +0.3900 | (11) -0.2510 | (12) -0.7250 | (13) -0.1625 | (14) +1.000 | (15) +0.2075 | (16) -0.0875 |
| (17) +0.4000 | (18) +0.5275 | (19) +0.3750 | (20) +0.2925 | (21) -0.5050 | (22) -1.000 | (23) -0.8400 | (24) -0.5775 |
| (25) -0.7125 | (26) +0.0500 | (27) +0.1700 | (28) +0.9450 | (29) -0.6200 | (30) +0.1500 | (31) 0.0000 | (32) +0.0400 |
| (33) -0.1000 | (34) +0.1600 | (35) +0.8050 | (36) -0.1000 | (37) +0.6000 | (38) -0.8050 | (39) +0.3325 | (40) +0.2675 |
| (41) +0.0300 | (42) +0.2000 | (43) +0.3100 | (44) -0.5200 | (45) +0.0950 | (46) -0.4000 | (47) +0.0525 | (48) -0.3800 |
| (49) +0.2000 | (50) -0.0500 | (51) -0.7500 | (52) -0.2350 | | | | |

Table 5: Week number (in brackets) and $E_{6,3,p}$ (p = 1,…,52).

| (1) +0.4725 | (2) +0.8750 | (3) +0.0975 | (4) +0.8275 | (5) +0.4000 | (6) +0.8100 | (7) +0.1500 | (8) +0.8825 |
|---|---|---|---|---|---|---|---|
| (9) +0.5400 | (10) +0.2500 | (11) +0.7675 | (12) +0.4025 | (13) +0.9175 | (14) +0.7850 | (15) +0.6350 | (16) +0.0700 |
| (17) +0.3375 | (18) -0.0675 | (19) +0.015 | (20) +0.1475 | (21) +0.8200 | (22) +0.7000 | (23) +0.3800 | (24) +0.0300 |
| (25) +0.6250 | (26) +0.0025 | (27) -0.1000 | (28) +0.2550 | (29) -0.0550 | (30) +0.1675 | (31) +0.3000 | (32) +0.1200 |
| (33) +0.1500 | (34) +0.2450 | (35) +0.8800 | (36) +0.4500 | (37) +0.6275 | (38) +0.1875 | (39) +0.9600 | (40) +0.2575 |
| (41) -0.0475 | (42) +0.3625 | (43) +0.0050 | (44) +0.2400 | (45) +1.000 | (46) +0.1425 | (47) +0.7050 | (48) +0.5825 |
| (49) -0.1600 | (50) +0.050 | (51) 0.0000 | (52) -0.0500 | | | | |

Table 6: Week number (in brackets) and $E_{8,6,p}$ (p = 1,…,52).

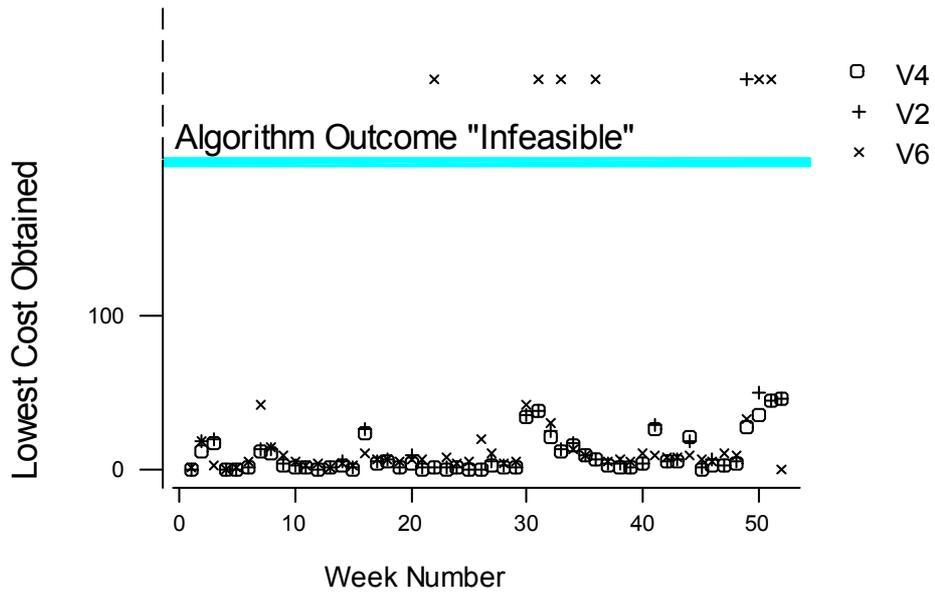

Figure 1: Comparison of the best solution found in 20 runs for all data instances for algorithms V2, V4 and V6.

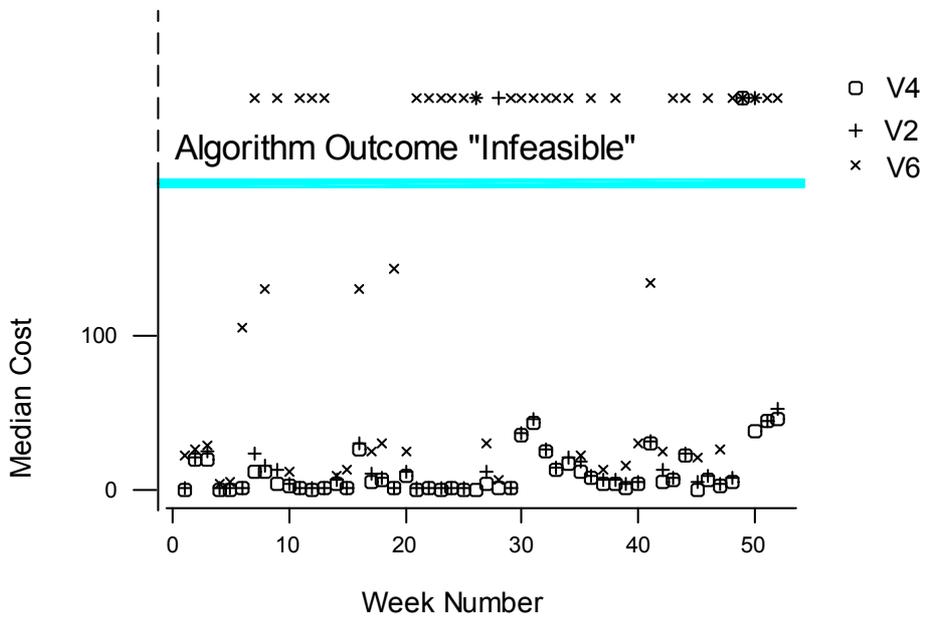

Figure 2: Comparison of the median solution found for all data instances for algorithms V2, V4 and V6.

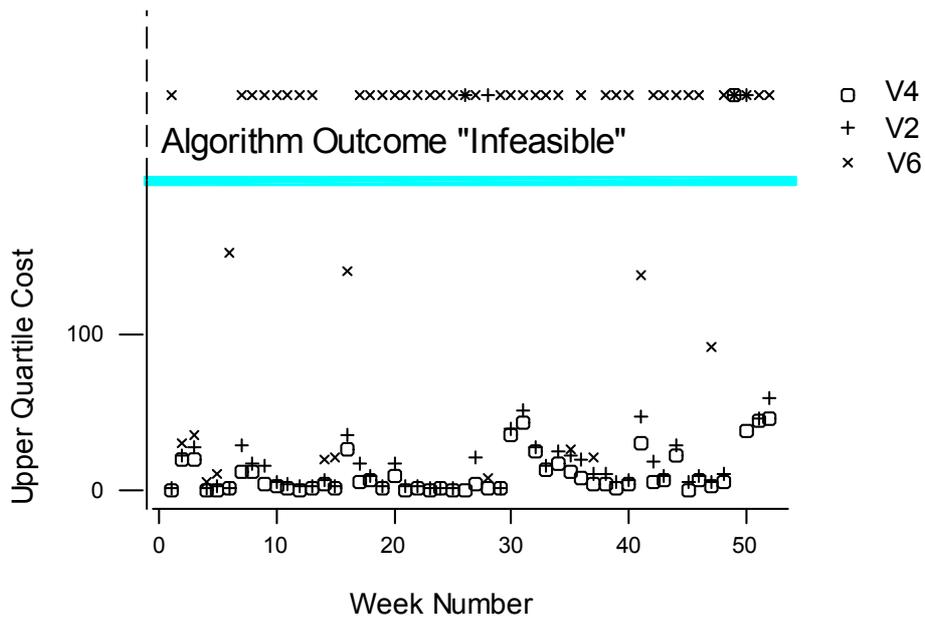

Figure 3: Comparison of the upper quartile of solutions for all data instances for algorithms V2, V4 and V6.

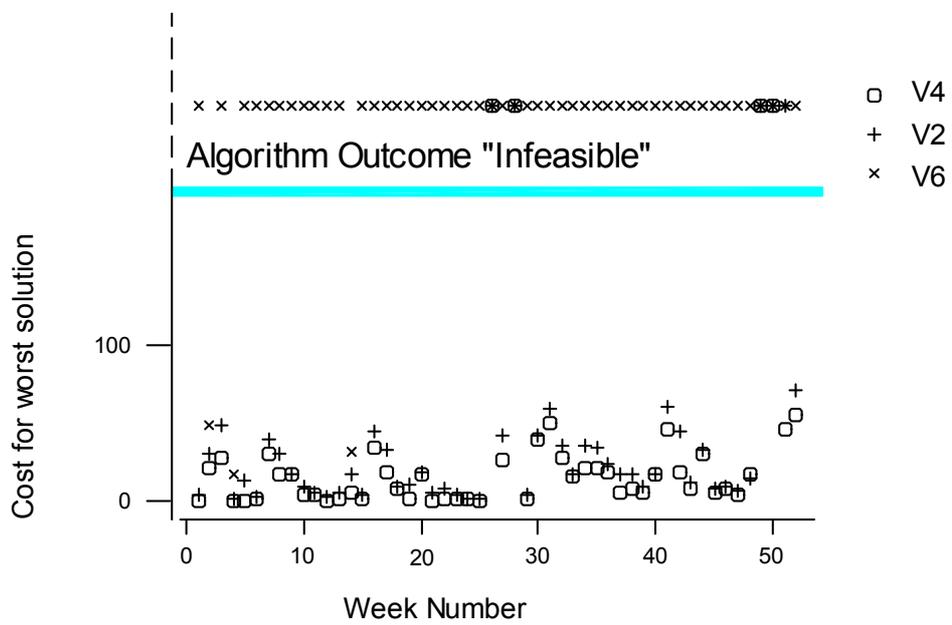

Figure 4: Comparison of the worst solution found in 20 runs for all data instances for algorithms V2, V4 and V6.

### Final Genetic Algorithm (U7)

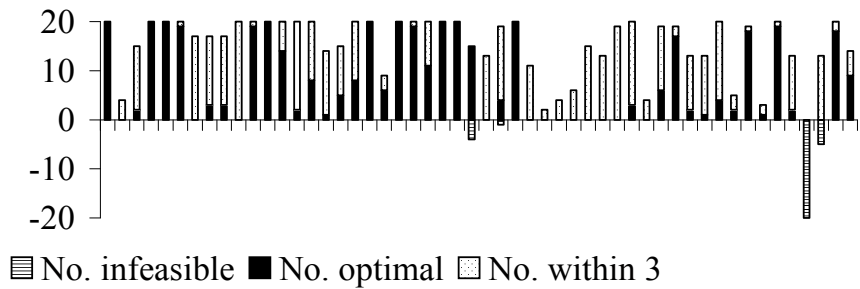

### Best Indirect Genetic Algorithm (V4)

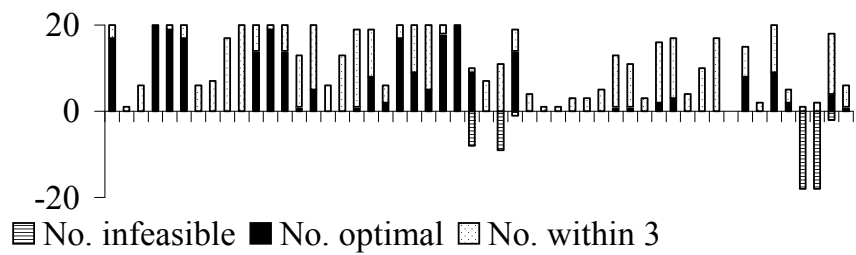

### Best Direct Genetic Algorithm (V7)

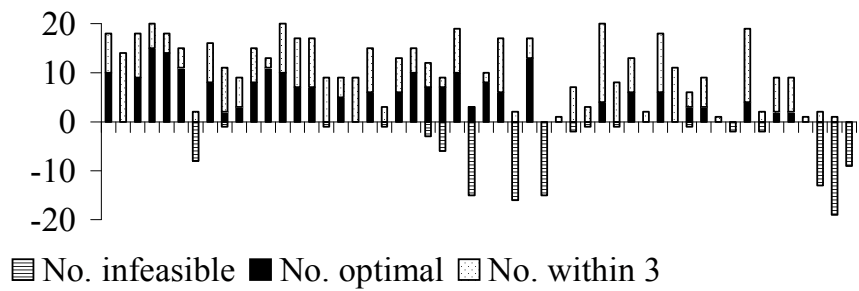

Figure 5: Summary results for all data sets. The bars show the number of infeasible, good feasible and optimal solutions out of 20, i.e. longer bars above the x-axis and shorter bars below the x-axis are better. Solutions within three units of optimality are shown as these are still acceptable to the hospital.